# Carbon Emissions and Large Neural Network Training

David Patterson[1,2], Joseph Gonzalez[2], Quoc Le[1], Chen Liang[1], Lluis-Miquel Munguia[1],
Daniel Rothchild[2], David So[1], Maud Texier[1], and Jeff Dean[1]
{davidpatterson, qvl, crazydonkey, llmunguia, davidso, maudt, jeff}@google.com,
{pattrsn, jegonzal, drothchild}@berkeley.edu

**Abstract:** The computation demand for machine learning (ML) has grown rapidly recently, which comes with a number of costs. Estimating the energy cost helps measure its environmental impact and finding greener strategies, yet it is challenging without detailed information.

We calculate the energy use and carbon footprint of several recent large models—T5, Meena, GShard, Switch Transformer, and GPT-3—and refine earlier estimates for the neural architecture search that found Evolved Transformer.

We highlight the following opportunities to improve energy efficiency and *$CO_2$ equivalent emissions* (*$CO_2$e*):
- Large but sparsely activated DNNs can consume <1/10th the energy of large, dense DNNs without sacrificing accuracy despite using as many or even more parameters.
- Geographic location matters for ML workload scheduling since the fraction of carbon-free energy and resulting $CO_2$e vary ~5X-10X, even within the same country and the same organization. We are now optimizing where and when large models are trained.
- Specific datacenter infrastructure matters, as Cloud datacenters can be ~1.4-2X more energy efficient than typical datacenters, and the ML-oriented accelerators inside them can be ~2-5X more effective than off-the-shelf systems.

Remarkably, the choice of DNN, datacenter, and processor can reduce the carbon footprint up to ~100-1000X.

These large factors also make retroactive estimates of energy cost difficult. To avoid miscalculations, we believe ML papers requiring large computational resources should make energy consumption and $CO_2$e explicit when practical. We are working to be more transparent about energy use and $CO_2$e in our future research. To help reduce the carbon footprint of ML, we believe energy usage and $CO_2$e should be a key metric in evaluating models, and we are collaborating with MLPerf developers to include energy usage during training and inference in this industry standard benchmark.

## 1. Introduction

As ML models increase in scale, a general trend is that they become more accurate and more capable. However, larger models translate to greater computing demands and, by extension, greater energy demands. We focus on natural language processing (NLP) because it is important in Google products and because of the recent development of many large NLP models, e.g., T5 [Raf19], Meena [Adi20], GShard [Lep20], Switch Transformer [Fed21], and GPT-3 [Bro20]. Recent studies attempt to evaluate the environmental impact of this trend in NLP, which is difficult [Str19]. Here we investigate and share the estimates of the energy consumed and $CO_2$e[3] of these recent and large NLP models. We also reduce by 88X an earlier estimate of the $CO_2$e for the neural architecture search for Evolved Transformer [So19, Str19] by characterizing the actual search process on the hardware and datacenter on which it was performed (see Appendices C and D).

Our investigation into $CO_2$e revealed surprises and misunderstandings about the full Deep Neural Network (DNN) lifecycle, the datacenters and hardware that run them, the variations in energy mix, and the difficulty of assessing $CO_2$e accurately. Note that we are evaluating the $CO_2$e of *operating* computers and datacenters, but not fabricating and recycling them (see [Gup20] for the latter topic).

To make it easier for the ML community to understand the real impact of training and how to reduce it, we endorse prior calls for new publication norms for computationally intensive ML models:

---

[1] Google
[2] University of California, Berkeley
[3] "$CO_2$e" means $CO_2$ *equivalent emissions*, accounting for carbon dioxide and all the other greenhouse gases as well: methane, nitrous oxide, ... (calculated from Equation A-1 in 40 Code of Federal Regulations 98). "$CO_2$ emissions" is only carbon dioxide. *t$CO_2$e* stands for 1000 kg (metric ton) of $CO_2$ equivalent emissions.

1. We must assess $CO_2e$ correctly, but it is hard to quantify precisely in part because all the required information is rarely reported or publicly available (e.g., datacenter, hardware, energy mix) and in part because it is hard to uncover important details afterwards (see Section 4.1). To make the carbon costs of training transparent, we encourage more researchers to measure energy usage and $CO_2e$—or to get a rough estimate using a tool like ML Emissions Calculator [Lac19] (Section 4.3)—and publish the data.
2. We agree with [Str19,Sch20,Hen20] that efficiency should be an evaluation criterion for publishing ML research on computationally intensive models besides accuracy and related measures, since we need to encourage advances across the board as the most sustainable energy is the energy you don't use.
3. And even if we could bring $CO_2e$ to zero in cloud datacenters, reducing training time matters, both because "time is money," and because cheaper training lets more people participate. Hence, we also second the recommendation of [Str19] for more researchers to publish the number of accelerators and their time to train computationally intensive models to inspire progress in reducing training costs.

We believe such new incentives could lead to a virtuous cycle where ML practitioners compete to increase accuracy while lowering energy consumption and $CO_2e$ that could bend the curve of ML carbon footprint growth for computationally intensive NLP models.

The following sections summarize the findings that led to these recommendations. They also document our $CO_2e$ estimates, highlight recent advances that curb the $CO_2e$ of ML, and estimate the $CO_2e$ from training the five recent large NLP models mentioned above. We end by updating the results of [Str19] on the emissions of the Evolved Transformer neural architecture search and discussing common misperceptions.

We start with an overview of the carbon footprint over the DNN lifecycle and show ways to improve a concrete example by nearly two orders of magnitude.

## 2. Energy Consumption and Carbon Footprint of an NLP Model

Electricity required to run an ML model is a function of the algorithm, the program that implements it, the number of processors that run the program, the speed and power of those processors, a datecenter's efficiency in delivering power and cooling the processors, and the energy supply mix (renewable, gas, coal, etc.). A simplified formula for the carbon footprint of an ML model that takes these factors into account is:

$$Footprint = (electrical\ energy_{train} + queries\ \times\ electrical\ energy_{inference}) \times CO2e_{datacenter}/KWh$$

Most companies spend more energy on serving a DNN model (performing inference) than on training it. For example, NVIDIA estimated that 80–90% of the ML workload is inference processing [Leo19]. Similarly, Amazon Web services claimed that 90% of the ML demand in the cloud is for inference [Bar19]. Given its substantial role in the ML model lifecycle, Alibaba, Amazon, Google, and NVIDIA designed ML accelerators solely for inference. If the total ML energy is split 10% on training and 90% on serving, then even if a given ML model required double the energy cost of training, it could reduce overall total carbon emissions if that model also cut serving energy by 20%. Because energy usage during training is more isolated and thus easier to investigate than inference, we focus on it in this paper, but keep in mind that the carbon footprint of inference is significant.

An ML practitioner is often improving the quality of an existing model rather than starting from scratch. We will use as a running example (found in [Str19]) the $CO_2e$ impact of going from training a Transformer model using off-the-shelf hardware in an average datacenter to training an Evolved Transformer model on Google's custom hardware for DNNs in Google's energy optimized datacenters. The large impact of each factor in this example demonstrates why we suggest that the trainers of a model be involved in the calculation of its costs.

Table 1 shows the $CO_2e$ breakdown, which we explain further in the next subsections along with the business rationale for these improvements, demonstrating the cross-cutting incentives for more efficient ML. Figure 1 illustrates the gains per step; the overall improvement in $CO_2e$ is 57X. This large gain demonstrates why the selection of the DNN model, processor, datacenter, and geographic location are critical to improve $CO_2e$. Table 2 shows the units for $CO_2e$ and a running example that puts these units into perspective.

We next go over the four factors in more detail that contribute to the carbon footprint of training.

| Model | Transformer (Big) | Evolved Transformer (Medium) | Transformer (Big) | Evolved Transformer (Medium) |
|---|---|---|---|---|
| Number of Parameters (B) | 0.21 | 0.13 | 0.21 | 0.13 |
| Datacenter | US Average | Google Iowa Council Bluffs | | |
| Datacenter Gross $CO_2$e/KWh (kg/KWh) 2020 (Section 2.4 and Appendix D) | 0.429 | 0.478 | | |
| Datacenter Net $CO_2$e/KWh (kg/KWh) 2020 (Section 2.4 and Appendix D) | 0.429 | 0.080 | | |
| Datacenter PUE (Latest quarter 2020) | 1.59 | 1.11 | | |
| Processor | P100 | | TPU v2 | |
| Chip Thermal Design Power (TDP in Watts) | 300 | | 280 | |
| Measured System Average Power including memory, network interface, fans, host CPU (Watts) | 296 | 271 | 229 | 227 |
| Measured Performance (TFLOPS/s)[5] | 6.7 | 4.7 | 28.8 | 24.0 |
| Number of Chips | 8 | | | |
| Training time to accuracy goal (days) | 3.5 | 3.2 | 0.81 | 0.62 |
| Total Computation (floating point operations) | 1.61E+19 | 1.03E+19 | 1.61E+19 | 1.03E+19 |
| Energy consumption (KWh) | 316 | 221 | 185 | 40 | 30 |
| Gross $CO_2$e for Model Training (metric ton) (Section 2.4 and Appendix D) | 0.1357 | 0.1055 | 0.0883 | 0.0189 | 0.0143 |
| Net $CO_2$e for Model Training (metric ton) (Section 2.4 and Appendix D) | 0.1357 | 0.0177 | 0.0148 | 0.0032 | 0.0024 |
| % 24/7 net carbon free energy (CY 2019) | N/A | 78% | | |

Table 1. See Appendix A for more detail[4]. Estimates of $CO_2$e for Transformer and Evolved Transformer for P100 and TPU v2 are based on power measurements.[5] Evolved Transformer (Medium) reached the same accuracy as Transformer (Big) in [So19]. $CO_2$e is shown both before ("gross") and after ("net") accounting for 24/7 reduction via real time, local carbon free energy purchases (Appendix B). To help put the $CO_2$e numbers in perspective, a single passenger round trip SF-NY is ~1.2t $CO_2$e (Table 2).

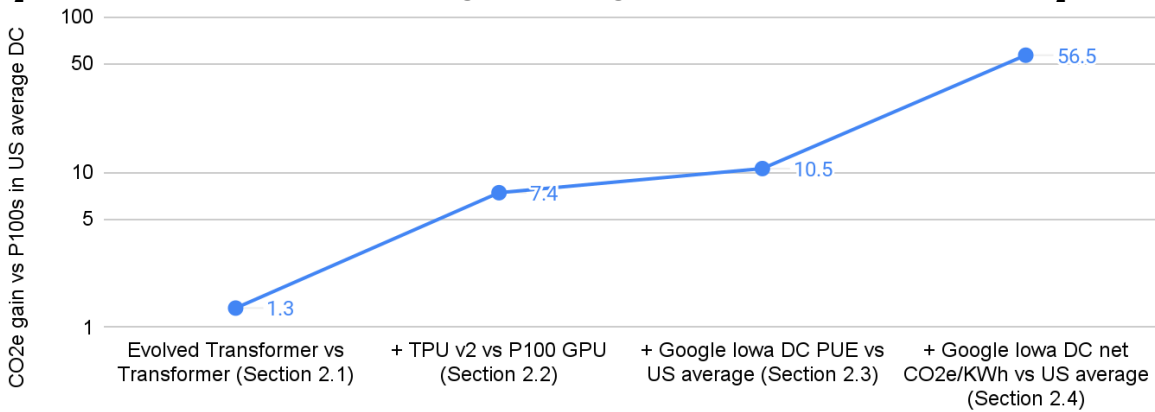

Figure 1. Improvement in $CO_2$e over Transformer (Big) on P100 GPUs in an average US datacenter versus Evolved Transformer (Medium) on TPU v2s in the Google Iowa datacenter.

| | Small Unit | Large Unit |
|---|---|---|
| Energy Consumption | Kilowatt hours (KWh) | Megawatt hours (MWh = 1000 KWh) |
| Carbon Footprint ($CO_2$e or $CO_2$) | Kilograms (kg) | Metric ton (t = 1000 kg) |
| Perspective (see Appendix A) | Single passenger round trip SF-NY (1.2t $CO_2$e) | Passenger jet plane round trip SF-NY (180t $CO_2$e) |

Table 2. Small and large units for energy and carbon footprint in this paper, plus airline travel $CO_2$e used for perspective on the relative size of ML emissions compared to other activities (Section 4.8).

---

[4] The peak TeraFLOPS/second is 19 for P100 and 46 for TPU v2.
[5] Training on TPU v3 instead of TPU v2 takes Transformer (Big) 0.44 days (averaging 61 TFLOPS/s) and 0.37 days (47 TFLOPS/s) for Evolved Transformer (Medium). For TPU v4, the respective numbers are 0.25 days (93 TFLOPS/s) and 0.19 days (73 TFLOPS/s). TPU v3 shrinks energy consumed and gross and net $CO_2$e from TPU v2 by ~1.4X for Transformer and by ~1.3X for Evolved Transformer.

## 2.1 Algorithm/program improvement

The Evolved Transformer (Medium) model discovered by So et al. [So19] using neural architecture search (see Section 4.1) uses 1.6X fewer FLOPS and 1.1X–1.3X less time than Transformer (Big) at slightly higher accuracy (see Table 1 and Appendix A)[6].

*Business Rationale*. Training faster saves ML researchers time as well as saves their organizations money and reduces $CO_2$e.

## 2.2 Processor improvement

Google's custom TPU v2 processor runs Transformer (Big) 4.3X faster than P100 GPUs and Evolved Transformer (Medium) 5.2X faster.[7] TPU v2 also uses less power: 1.3X less for Transformer and 1.2X less for Evolved Transformer. The net gain in performance/Watt is 5.6X and 6.2X, respectively.

*Business Rationale*. The substantial increase in the scope and scale of deep learning over the past decade has created the opportunity to build customized hardware that is tailored to the kinds of computations involved in training and serving DNN models. Instead of using GPUs like many other organizations, over the past seven years Google has designed, built, and deployed four generations of custom Tensor Processing Unit (TPU) hardware for DNNs to accelerate model training and serving [Jou21]. To get a better return on their investment, cloud companies actually aim for improved *cost-performance,* as opposed to simply performance. Cost here means *Total Cost of Ownership* (*TCO*), which includes the annual operating costs such as electricity consumed and amortization of capital expenditures for the computer, cooling, power distribution, and the building. Jouppi *et al*. show that power consumption is nearly perfectly linearly correlated with TCO[8] [Jou21], so performance/TCO gains also help performance/Watt, saving money and reducing $CO_2$e.

## 2.3 Datacenter improvement

A useful quantitative metric of datacenter efficiency is the energy overhead above and beyond what directly powers the computing equipment inside the datacenters. If the overhead were 50%, the *Power Usage Effectiveness* (*PUE*) would be 1.50. The US national datacenter average in 2018 was 1.58, which is the value [Str19] used; In 2020, it was 1.59. Google publishes its datacenter PUE online every quarter. The PUE for the Iowa datacenter where we ran Evolved Transformer is 1.11, a factor of 1.4X better. Cloud datacenters are roughly 2X as energy efficient as a typical enterprise datacenter due to other factors like server utilization (see [Höl20]), but we'll limit the quantitative improvement in this paper to the easy-to-measure PUE.

More broadly, since cloud datacenters are much more energy efficient, the long-feared explosion of datacenter energy usage has not materialized. A recent paper in *Science* [Mas20] found that global datacenter energy consumption increased by only 6% compared with 2010, despite computing capacity increasing by 550% over the same time period [Mas21].

*Business Rationale*. Cloud companies strive for energy efficient datacenters since it saves money and lowers emissions. Perhaps we should add "energy is money" to Ben Franklin's "time is money" advice?

## 2.4 Energy mix improvement

The gross carbon intensity of energy according to the U.S. average mix is 0.429 kg of $CO_2$e/KWh [USE21]. After matching Google's clean energy purchase per its 24/7 carbon-free energy framework (see Appendix B), the net $CO_2$e drops to 0.080 for the Iowa datacenter where we ran Evolved Transformer, which is 5.4X better.

*Business Rationale*. Transmitting electricity long distances is more expensive and less efficient than sending information as photons over optical fibers [Arm10]. Cloud computing allows companies like Google to have a global portfolio of datacenters, many of which are placed where the grid is cleaner (e.g., Finland) or where companies can purchase clean energy directly (e.g., Iowa). In 2020 Google announced a new objective in its energy strategy: by 2030, it aims to run all Google datacenters and offices on carbon-free energy 24/7. For our 24/7 carbon-free energy accounting (see Appendix B), we deduct from the hourly consumption all clean energy purchased on that same geographically local grid and the same hour, which results in the net $CO_2$e/KWh value. As Iowa has strong nighttime winds, Google's wind portfolio lowered Iowa's datacenter *gross* average $CO_2$e/KWh in December 2020 by 6X, from the local grid's 0.478 kg to a *net* average of 0.080 kg.

---

[6] Their neural architecture search also found another version that had the same performance but better accuracy.
[7] [Str19] used P100s, which are contemporary GPUs to TPU v2s.
[8] The correlation coefficient R between TCO and TDP is 0.99 out of 1.00 across four generations of TPUs.

## 2.5 Summary: Formulas for energy consumption and carbon footprint of training

Reducing $CO_2$e is not only a moral obligation but ultimately sound business. To decrease the footprint of training, an ML researcher should pick the DNN model, the processor, and the datacenter carefully.[9] Cutting energy saves money and $CO_2$e and improving the energy mix reduces $CO_2$e. We refactor the equation above for training into energy consumption and its carbon footprint (t$CO_2$e means metric tons of $CO_2$e):

$$KWh = Hours\ to\ train \times Number\ of\ Processors \times Average\ Power\ per\ Processor \times PUE \div 1000$$

$$tCO2e = KWh \times kg\ CO2e\ per\ KWh \div 1000$$

We believe it is straightforward for ML practitioners to calculate energy consumption. They already know hours to train and number of processors. Google and Facebook publish PUE of their datacenters, so that is easy to look up for those clouds. If cloud providers don't share PUE, use the US average PUE as in [Str19]. We measured the power of the processors during training, which is ideal, but using the average of the training of several similar models is probably sufficient and much easier.[10] Table 3 shows the average power and standard deviation for the processors and DNNs that we measured in this paper.

The final piece is the $CO_2$e of the datacenter at the time the model was run. Google calculates the average per month, which is close enough, and it is now available for Google employees to look up. Without access to such a dashboard, use the ML Emissions Calculator [Lac19] or Green Algorithms tool [Lan20] that estimate the $CO_2$e mix by region (see Figure 6 below)[11]. While not absolutely necessary, we hope the ML community will lobby all cloud providers to reveal the actual energy mix, since it can vary within a region. For example, to let customers pick the datacenter based on $CO_2$e, Google Cloud recently released the percentage of carbon-free energy and gross $CO_2$e of its datacenters and committed to publishing updated figures going forward.

We next show the impact of these three choices on much larger NLP models.

| Processor | Average (Watts) | StDev % | DNNs used to calculate average power |
|---|---|---|---|
| TPU v2 | 221 | 5% | Transformer (Big), Evolved Transformer (Medium), Neural Architecture Search [So19] |
| TPU v3 | 283 | 10% | T5, Meena, Gshard, Switch Transformer |
| P100 GPU | 271 | 11% | Transformer (Big), Evolved Transformer (Medium), Neural Architecture Search [So19] |
| V100 GPU | 325 | 2% | Transformer (Big), GPT-3 [Sut21] |

Table 3. Average system power per processor and standard deviation for DNNs in this paper. We measured the Google DNNs (see Tables 1 and 4). OpenAI measured GPT-3 in a Microsoft Azure datacenter [Sut21].

## 3. Energy Usage and $CO_2$e Emissions of Five Recent Large NLP Models

A natural question that follows is what about the training $CO_2$e of much larger NLP models? Table 4 and Appendix A show a $CO_2$e calculation[11] for five of them: T5, Meena, GShard, and Switch Transformer from Google plus GPT-3 from Open AI that runs on Microsoft Azure Cloud:

- *T5* is a pre-trained language model that casts all NLP problems in a unified text-to-text format to enable application of transfer learning techniques to reduce the cost of training [Raf19]. The largest size has 11B parameters, and training used 86 MWh and produced 47 t$CO_2$e.
- *Meena* is a multi-turn open-domain chatbot [Adi20]. This 2.6B parameter DNN is trained to minimize perplexity of the next token. The year-old companion paper has ~150 citations. Training Meena used

---

[9] PUE and kg $CO_2$e per KWh are functions of the datacenter where the model is run.
[10] The ML Emissions Calculator [Lac19] also estimates power per processor. It now uses the values in Table 3 for TPU v2 and TPU v3 [Luc21]. At the time of this writing, the calculator shows $CO_2$e produced but not the estimated power per processor, energy consumed, or $CO_2$e/KWh.
[11] The Google models happen to be run in datacenters where the gross and net $CO_2$e were the same or close.

232 MWh and emissions was 96 tCO₂e. As Evolved Transformer saved 48 tCO₂e alone for the single use case of developing Meena (see Table 4), the 3.2 net tCO₂e cost for its development returned 15:1.
- *GShard* is composed of a set of lightweight annotation APIs that provide an elegant way to express a wide range of parallel computation patterns with minimal changes to the existing model code [Lep20]. It enabled scaling up of a multilingual neural machine translation Transformer model with sparsely gated mixture-of-experts (MoE) [Sha17] using automatic sharding. The GShard-600B model is a particular use of that framework for training a multi-lingual translation model with 600B total parameters. Sparse models can have many model parameters while requiring much less computation than dense models. Training GShard-600B used 24 MWh and produced 4.3 net tCO₂e.
- *Switch Transformer* simplifies the Mixture of Expert (MoE) routing algorithm to design intuitive improved models with reduced communication and computational costs [Fed21]. The authors show large sparse models—1500B parameters but only 0.1% activated per token—can deliver up to 7x increases in pre-training speed with the same computational resources. We estimated it used 179 MWh and produced 59 net tCO₂e.

| Model | Evolved Transformer NAS | T5 | Meena | Gshard -600B | Switch Transformer | GPT-3 |
|---|---|---|---|---|---|---|
| Number of Parameters (B) | 0.064 per model | 11 | 2.6 | 619 | 1500 | 175 |
| Percent of model activated on every token | 100% | 100% | 100% | 0.25% | 0.10% | 100% |
| Developer | | | Google | | | OpenAI |
| Datacenter of original experiment | Google Georgia | Google Taiwan | Google Georgia | Google North Carolina | Google Georgia | Microsoft |
| When model ran | Dec 2018 | Sep 2019 | Dec 2019 | Apr 2020 | Oct 2020 | 2020 |
| Datacenter Gross CO₂e/KWh (kg/KWh when it was run) | 0.431 | 0.545 | 0.415 | 0.201 | 0.403 | 0.429 |
| Datacenter Net CO2e/KWh (kg/KWh when it was run) | 0.431 | 0.545 | 0.415 | 0.177 | 0.330 | 0.429 |
| Datacenter PUE (when it was run) | 1.10 | 1.12 | 1.09 | 1.09 | 1.10 | 1.10 |
| Processor | TPU v2 | | TPU v3 | | | V100 |
| Chip Thermal Design Power (TDP in Watts) | 280 | | 450 | | | 300 |
| Measured System Average Power per Accelerator, including memory, network interface, fans, host CPU (W) | 208 | 310 | 289 | 288 | 245 | 330 |
| Measured Performance (TFLOPS/s)[12] | 24.8 | 45.6 | 42.3 | 48.0 | 34.4 | 24.6 |
| Number of Chips | 200 | 512 | 1024 | 1024 | 1024 | 10,000 |
| Training time (days) | 6.8 | 20 | 30 | 3.1 | 27 | 14.8 |
| Total Computation (floating point operations) | 2.91E+21 | 4.05E+22 | 1.12E+23 | 1.33E+22 | 8.22E+22 | 3.14E+23 |
| Energy Consumption (MWh) | 7.5 | 85.7 | 232 | 24.1 | 179 | 1,287 |
| % of Google 2019 total energy consumption (12.2 TWh = 12,200,000 MWh) [Goo20] | 0.00006% | 0.00070% | 0.00190% | 0.00020% | 0.00147% | 0.01055% |
| Gross tCO₂e for Model Training | 3.2 | 46.7 | 96.4 | 4.8 | 72.2 | 552.1 |
| Net tCO₂e for Model Training | 3.2 | 46.7 | 96.4 | 4.3 | 59.1 | 552.1 |
| Fraction of NAS Estimate in [Str19] (284 tCO2e) | 0.011 | 0.164 | 0.340 | 0.015 | 0.208 | 1.944 |
| Fraction of equivalent jet plane CO₂e round trip San Francisco ↔ New York (~180 t; see Ap. A) | 0.018 | 0.258 | 0.533 | 0.024 | 0.327 | 3.054 |
| tCO₂e savings by Meena using Evolved Transformer | -- | -- | 48.5 | -- | -- | -- |
| % 24/x7 carbon free energy (when run) | 31% | 19% | 30% | 73% | 43% | N/A |

Table 4. CO₂e for NLP models (see Appendix A)[12]. V100's TDP is closer to average power due to Turbo mode and DVFS. TPUs don't offer them, so their TDP is much higher than their average power.

---

[12] The peak TeraFLOPS/second is 46 for TPU v2, 123 for TPU v3, and 125 for V100.

- *GPT-3* is an autoregressive language model with 175B parameters, 10x more than any non-sparse language model at the time [Bro20]. It achieves strong performance on many NLP datasets. A winner of the best paper award at NeurIPS 2020, this 8-month-old paper already has ~700 citations and made mainstream media headlines.[13] It is now available for commercial use. One potential energy benefit of a large language model like GPT-3 is that they exhibit few-shot generalization, which means that they don't need to be retrained for every new task like smaller models [Wan20]. Its estimated carbon emissions due to training are 552 $tCO_2e$ and its energy consumption is 1287 MWh.[14]

Table 4 also lists the neural architecture search for Evolved Transformer, discussed shortly.

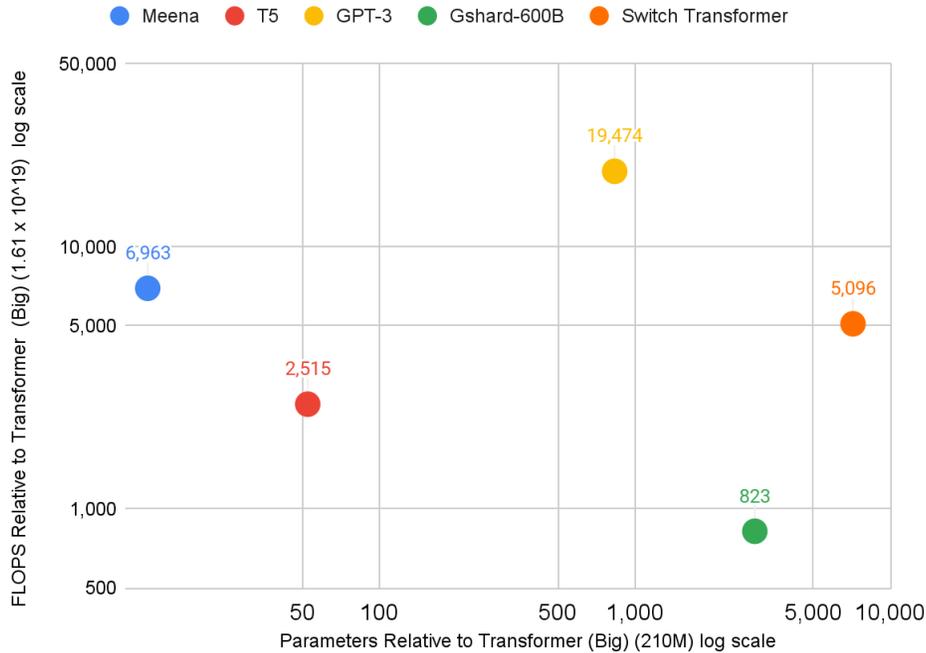

**Figure 2. Total FLOPS versus number of parameters relative to Transformer (Big) in a log-log graph (Table 1). While all are not doing the same tasks, a reason T5 has relatively lower FLOPS relative to its number of parameters is that it trains until the accuracy is good enough instead of to the best possible accuracy. [Kap20] notes that some architectures have a much lower footprint than others at equivalent accuracy and suggests that significant power might be saved by revisiting accuracy requirements.**

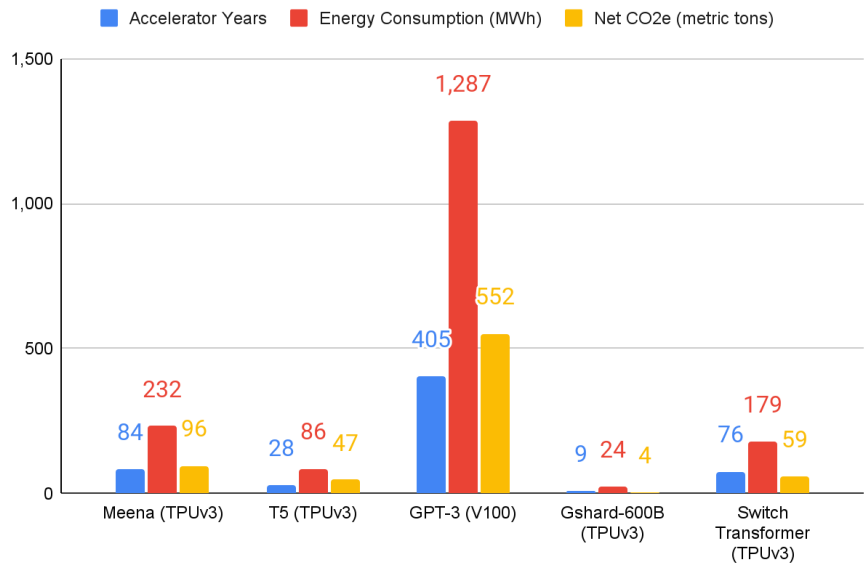

**Figure 3. Accelerator years of computation, energy consumption, and $CO_2e$ for five large NLP DNNs.**

---

[13] Metz, C., Meet GPT-3. It Has Learned to Code (and Blog and Argue), November 24, 2020, *New York Times*.
[14] We measured all the data for Google models. OpenAI measured V100 performance, V100 power, total FLOPS, and PUE for GPT-3. We used the US average $CO_2e$/KWh for GPT-3 at Microsoft Azure (see Appendix A).

Figures 2 and 3 present the same data graphically. Figure 2 plots the number of parameters on the X axis and number of total FLOPS on the Y axis relative to Transformer (Big) [So19] using a log-log graph. Sparsely activated models use many more parameters with much lower total FLOPS. Since performance is not necessarily linear in FLOPS (see [Li21]), Figure 3 shows computation in processor years along with their energy consumption and carbon footprint. Compared to the dense GPT-3, sparsely activated Gshard needs ~45X fewer processor years, uses ~55X less energy, and reduces gross $CO_2e$ ~115X and net $CO_2e$ ~130X.

## 4. Discussion

In this section, we address the additional factors relating to carbon emissions due to training NLP models. We start by revisiting the estimate of neural architecture search in [Str19] and end with example benefits of some NLP models.

### 4.1 Estimating the cost of neural architecture search (NAS)

The Evolved Transformer neural architecture search (NAS) was used as an example of an expensive NLP model [Str19]. Although it is now surpassed by other models in terms of training cost (Table 4), we discuss it here as a concrete example of the complexity of estimating the cost of a ML method retroactively.

As Table 4 shows, the actual cost of Evolved Transformer NAS is nearly two orders of magnitude smaller than previously estimated [Str19]. Why the discrepancy? The answer is that, in addition to the efficiency of Google datacenters, there was a confusion in estimating the energy cost of NAS. In Evolved Transformer NAS, researchers used a small *proxy task* to search for the best models to save time and money, and then scaled up the found models to full size. Small proxies may not be obvious, which made it hard to estimate the $CO_2e$ correctly in retrospect from the NAS paper [So19]. Due to the misunderstanding of the usage of proxy tasks in NAS, it was assumed the search was done with full size tasks. Because of this assumption, despite considerable effort on their part, Strubell *et al.*'s energy estimate for NAS ended up 18.7X too high for the average organization (see Appendix C) and 88X off in emissions for energy-efficient organizations like Google (see Appendix D). This example led us to our first recommendation—that more researchers measure energy usage and $CO_2e$ for computationally intensive projects, and report them when practical, rather than counting on others to estimate it retrospectively.

Another confusion in the general public is the misperception that NAS (and therefore, the cost associated with NAS) is conducted once per model training. In practice, however, NAS is generally not performed once per model training, but once per *problem domain+architectural search space combination*. For example, the Evolved Transformer, found by NAS on translation, can be used for language modeling without a new search [So19, Adi20]. Unfortunately, results in the earlier work by [Str19] characterizing NAS were misattributed to single model training costs in the popular press.

As an analogy, NAS is like optimizing the energy efficiency and cost of an LED light bulb with extensive simulations on a supercomputer, training a model is akin to building LED light bulbs, and inference is analogous to all the customers using LEDs to light their homes. The analogous confusion would be claiming that the one-time upfront supercomputer simulation cost should be included in the $CO_2e$ cost of every light bulb manufactured. In this analogy, the onetime $CO_2$ expenditure of the supercomputer simulations can be more than paid back with the improved energy-efficiency of the mass-produced light bulbs, as was the case for the actual NAS of [So19] (see next paragraph).

In terms of cost-benefit tradeoff, NAS can also lead to improved energy efficiency in training of downstream applications, and the benefit can dramatically outweigh the cost. Figure 4 shows that the Evolved Transformer, found by NAS [So19], has 37% fewer parameters and converges to the same accuracy with 25% less energy expenditure (see Table 1) than the vanilla Transformer (Big) model on WMT English to German translation. The use of Evolved Transformer instead of a regular Transformer architecture saved 48.5 $tCO_2e$ during the training of the Meena DNN (see Tables 1 and 4). The savings from this single reuse in Meena are ~15X larger than the energy cost of running the search to discover it. The results of the Evolved Transformer neural architecture search have been open-sourced. It can readily be used by anyone training ML models for NLP problems, similar to how a Transformer-style model can be used for NLP problems [Evo19].[15]

It would be beneficial to compare the cost-savings ratio of the Evolved Transformer NAS to previous work developing more efficient architectures. Unfortunately, as others have pointed out [Dod19, Str19], the full cost of model development is rarely, if ever, reported in the literature, making it impossible to compare this analysis to prior work, and preventing straightforward comparison among different approaches more generally.

This lack of training development costs is one example of how adopting higher standards for measuring and reporting ML model energy requirements would lead to a better understanding of cost-accuracy tradeoffs in ML models, potentially further reducing overall emissions by empowering more informed ML model selection, as the next subsection explains.

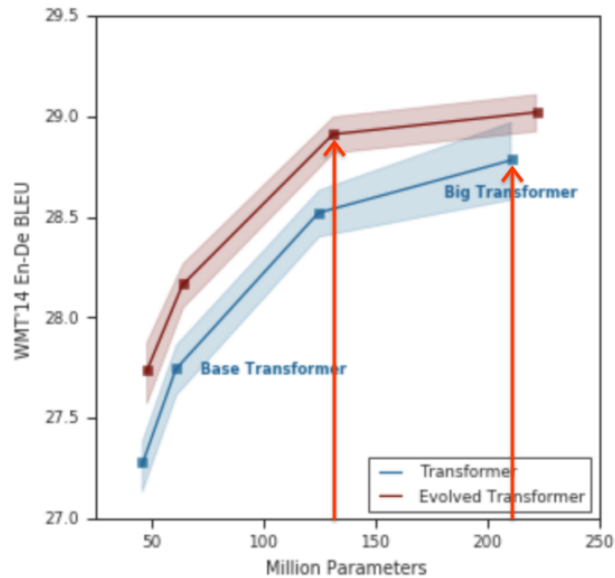

**Figure 4: Reproduction of Figure 4 from So *et al.* Dots on the blue line represent various sizes of plain Transformer NLP models, while dots on the red line represent various sizes of the open-sourced Evolved Transformer architecture that was discovered by the neural architecture search run in [So19]. Red arrows are at 131M and 210M parameters and show that an Evolved Transformer can achieve higher accuracy at less cost: it runs 1.3X faster and produces 1.3x less $CO_2$e.**

### 4.2 There are more resources used for training than the only final training run

[Str19] and others point out that it often takes many attempts to get everything set up correctly before the final training run, so the final training run does not reflect the total cost. Since it's hard to improve what you can't measure, one issue is how to account for such costs accurately. Fortunately, an internal Google product is underway that will record information about the training process, originally intended to keep track of information like data provenance. The developers now plan to add energy consumption so that Googlers can better understand the full training lifecycle. An example of an open source tool to record such information is experiment-impact-tracker [Hen20]. In addition, the developers of ML Emissions Calculator [Lac19] are currently working on CodeCarbon, whose goal is to measure/approximate carbon consumption automatically.

Alas, there will be no way to verify the claims in papers of preliminary training development. A lesson of computer benchmarking is that requiring the release of all information so that others could recreate your results was an effective deterrent to fudging the numbers. If more computationally intensive ML papers included energy consumption and carbon footprint of the final training run with sufficient details that others could check, that would be a great step forward. Perhaps ML practitioners could study the total lifecycle to develop rules of thumb to estimate the overall carbon footprint based on its final training cost.[16]

---

[15] Reuse reduces overall development effort and energy usage. For example, implementations of EfficientNets, EfficientDets [Tan19], developed via NAS for image-classification and object-detection, were forked on GitHub >4000 times.

The next subsection also emphasizes the value of measurement.

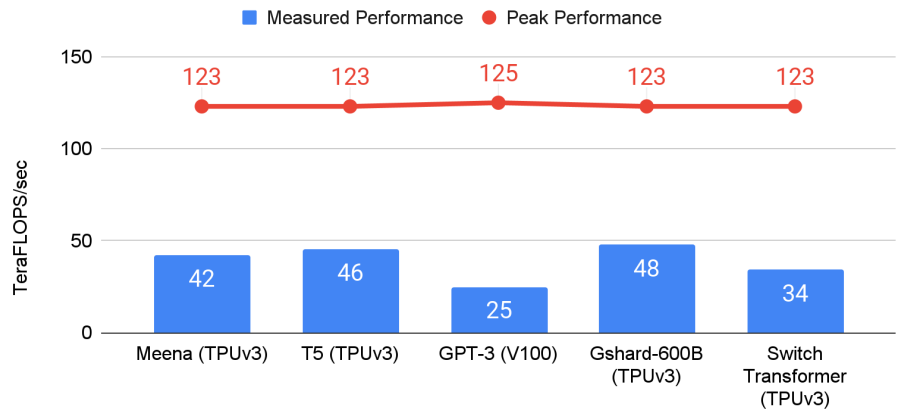

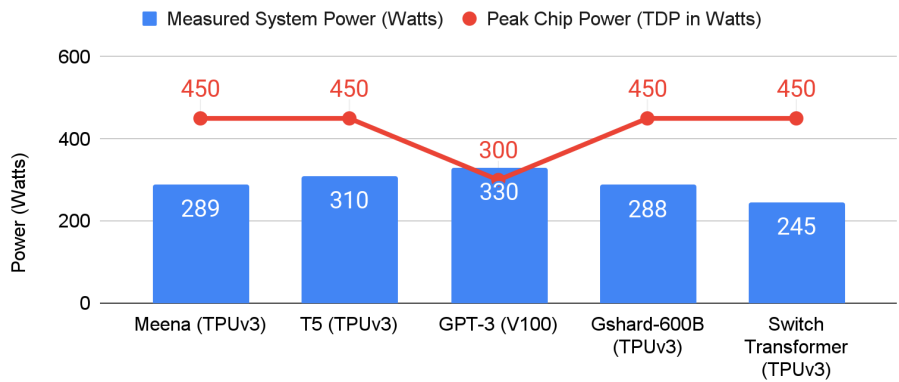

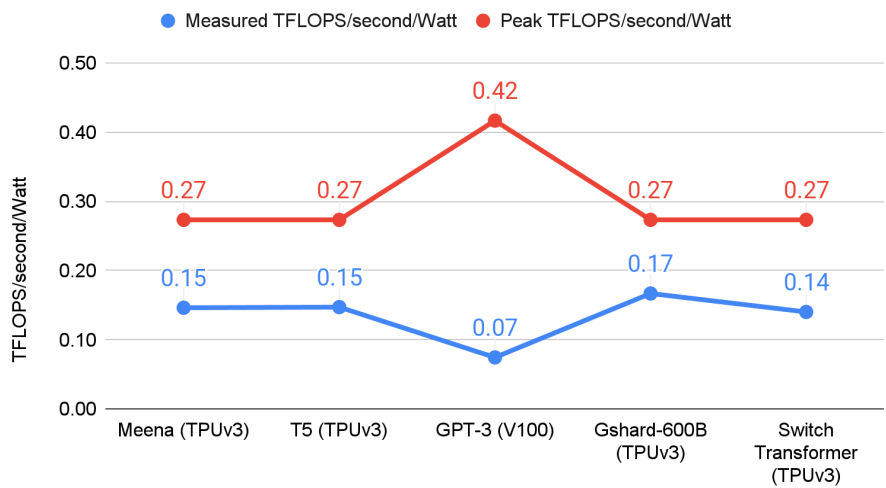

**Figure 5. Measured vs peak performance, measured system power vs peak chip power (TDP), and measured vs peak performance/Watt for V100 GPU and TPU v3 (see Table 4 and Appendix A).**

### 4.3 Measurements are more interesting than extrapolations

Although extrapolations of carbon emissions are relatively easy, more attention should be paid to actual experiments that have been conducted rather than to hypothetical case studies. As a problematic example, let's hypothesize what the $CO_2e$ would be for training Transformer (Big) on the CTS-1 Quartz - Tundra Extreme Scale supercomputer at Lawrence Livermore National Laboratory, one of the top 500 supercomputers (but one whose design is not optimized for ML training). Its ~100,000 cores might use ~75 MWh of power and might generate 32 $tCO_2e$, ~10,000 times larger than for TPU v2s at Google (Table 1)[17].

---
[16] Since large NLP models can take a month to train, developers cannot afford to do the full training task many times. Like [So19] for NAS, they likely use a smaller task to explore the space for a limited training time. One indication comes from the AutoML work in [Li21]. Their exploration computation cost was roughly equal to the final training cost.

The measurement advice applies to processors as well DNNs. Tables 1 and 2 show that the theoretical performance per Watt is higher than the measured performance per Watt on average by factors of 1.6X for TPUs and by 3.5X for GPUs. Figure 5 shows the information in Table 1 graphically. Using theoretical performance per Watt, V100 is 1.5X better than TPU v3, but it's the other way around for measured performance per Watt: TPU v3 is 2.0X better than V100 on average for these large NLP DNNs.

Figure 6 compares the gross $CO_2e$ estimates from the ML Emissions [Lac19] and Green Algorithms [Lan20] calculators to the processors and programs in this paper at the time of this writing (April 2021). Compared to the results in Tables 1 and 4, they differ by factors of 0.53–1.64 and 0.91–2.42 with geometric means of 0.92 and 1.48, respectively[18]. **The ML Emissions and Green Algorithms calculators do not estimate net $CO_2e$, which could be up to 10X lower.** The figure once again shows the increase in accuracy of measurement over indirect calculations. The authors of the Emissions Calculator agree that measurement is preferred, with some calculator as the best alternative if measurement is difficult to perform [Luc21].

The next discussion topic reminds us that improving the algorithm is often more important than improving the hardware.

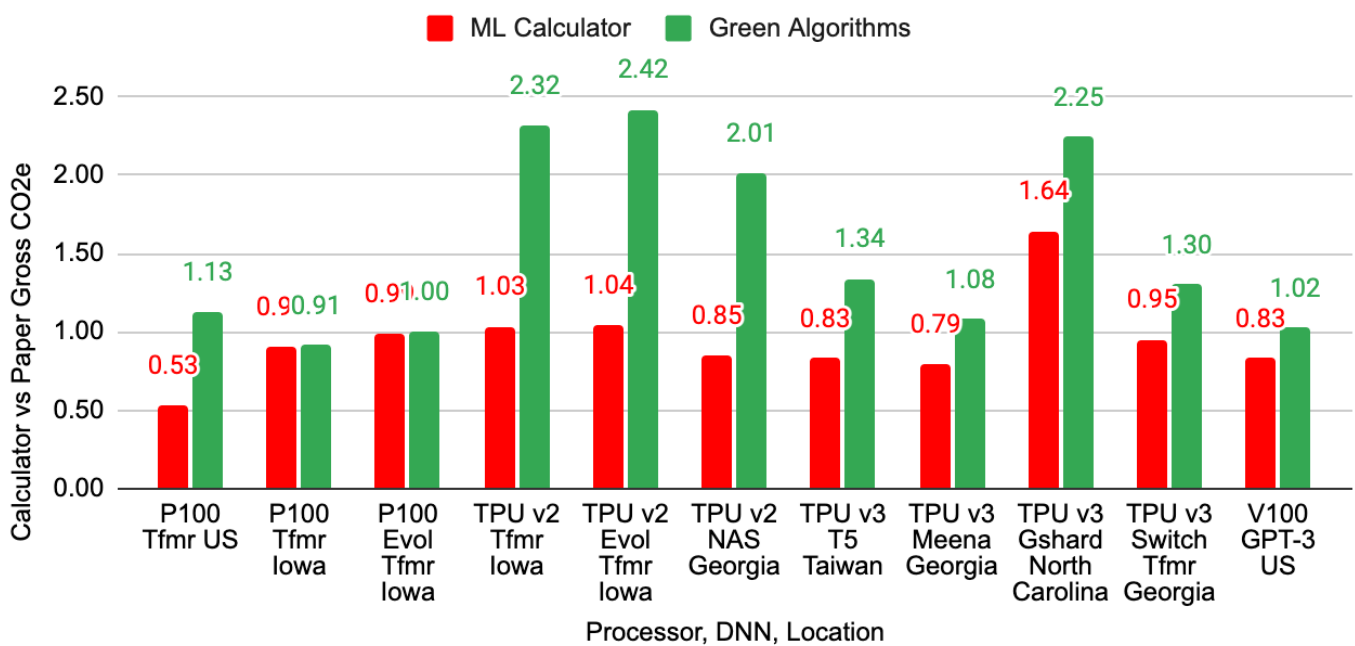

**Figure 6. Ratio of ML Emissions and Green Algorithm calculators vs gross $CO_2e$ in Tables 1 and 4.**

### 4.4 Standard ML algorithmic techniques can improve energy efficiency

There are many algorithmic techniques that can improve the energy efficiency of machine learning models. Some techniques can achieve the same accuracy with less overall computation. Others can use a large, already-trained model as a starting point and yield a lighter-weight, more computationally efficient model with almost the same accuracy. These techniques all serve to reduce the computational cost and therefore energy and carbon emissions of models. Some of these techniques include:
- *Distillation* transfers the knowledge from large models into smaller, more computationally efficient models [Hin15, San20].
- *Pruning*, *quantization*, and *efficient coding* can improve the energy efficiency of DNNs 3X–7X [Han15].

---
[17] We use US averages for kg $CO_2e$/KWh and datacenter PUE and assume it runs at 40% of the peak floating point performance of Quartz-Tundra (3.2 PetaFLOPS/sec). For reference, Figure 5 shows V100 running at 20% of peak.
[18] We picked the closest geographic option per calculator to the actual location in each case. The Green Algorithms paper lists Meena $CO_2e$ as 164t [Lan20], but the calculator result as of April 2020 was 85t for Virgina using Google Cloud.

- *Fine-tuning* and *transfer learning* both reuse already-trained representations, rather than starting training of each NLP task's parameters from random initialization, for example [Dod20].
- *Sparsely activated mixture-of-expert-style models* can provide more than 10X reductions in computation requirements and energy costs for both training and inference while providing significantly higher accuracy than dense Transformer or LSTM-based models of equivalent computational cost per token [Sha17,Lep20,Fed21]. Gshard-600B is one example, evaluated in Section 3.

We commend the development of such techniques. Some publication venues, such as the EACL and NAACL 2021 NLP conferences, have begun specifically soliciting research of this nature by offering "Efficient and Green" research tracks, alongside workshops such as SustaiNLP and EfficientQA. We encourage other venues to follow suit, and hope that many researchers will consider this line of work.

The next topic discusses one of our biggest surprises of this investigation, the importance of geography.

## 4.5 It matters which datacenter is used, even within the same organization

We were amazed by how much it matters *where* and *when* a DNN is trained. Moreover, this option is likely the easiest path for ML practitioners to reduce $CO_2$e. For example, after reading early drafts of this paper, some colleagues switched to a Google datacenter with a smaller carbon footprint to train a large NLP model.

Reviewers of early drafts suggested that datacenter energy use is a zero-sum game. They thought that any tasks run in a green datacenter simply shift other work to dirtier datacenters, so there is no net gain. It's not true, but that speculation reveals many seemingly plausible but incorrect fallacies:

- *Fallacy: Datacenters are fully utilized*. Applications are deployed to handle worst case demand depending on the time of day and day of the week, so for much of the time resources are idle [Arm10].
- *Fallacy: Cloud centers can't grow*. Similar to the founding of a new university, cloud companies buy much more land than they need initially at a site so that they can construct more buildings in the future without first traversing the lengthy process of acquiring land [Bar18].
- *Fallacy: Renewable energy is fixed and can't grow*. There is often an excess of renewable energy at some times of day (see Appendix B). The amount of solar and wind energy is also a function of the investment as well as weather conditions. Google's long term renewable energy procurement normally invests in the creation of new renewable energy resources. The greater the use and investment in renewable energy, the more money is available to buy and deploy new solar panels and wind turbines, thereby increasing the renewable energy supply. Thus, it's *not* the case that Google's use of renewable energy means other residents must use dirty energy. Appendix B introduces issues around carbon free energy use and investment.
- *Fallacy: Google NLP model training competes with other tasks in the datacenter*. Google trains large models on ML supercomputers that even have their own interconnection network, so ML training is distinct from CPU-only tasks [Jou20]. Tasks for CPUs don't interfere with TPUs, and vice versa.
- *Fallacy: Training must run in all datacenters*. While user facing inference applications need global distribution in order to provide low-latency access to users all around the world [Jou21], there is no problem to limit ML training computation to a smaller number of (green) datacenters. For example, Google is currently deploying numerous TPU v4s, many of which will be located in windy Oklahoma, whose net $CO_2$e/KWh is even lower than Iowa.
- *Fallacy: There is no business reason to reduce carbon emissions*. Reducing climate change certainly has long-term economic benefits for everyone. Google has been carbon neutral since 2007 and has procured enough additional renewable energy to match 100% of its datacenter energy usage since 2017, so the impact of the remaining carbon from training at Google is zero even today. Other hyperscalers aim for carbon neutrality by 2025 or 2030, so the whole cloud may become carbon neutral. With its new 24/7 local carbon-free energy goal by 2030, Google is now focused on purchasing carbon-free energy to match its hourly load at the same location as its datacenters with the goal to decarbonize its electricity supply (see Appendix B).

The next question that arose is whether such green datacenters are available to only a few ML practitioners.

### 4.6 Many have access to energy-optimized datacenters

The increasing use of cloud computing has decreased the energy intensity[19] of datacenters 20% annually since 2010 [Has20]. Access to energy-optimized, low-cost cloud datacenters is not restricted to employees of a few companies; people around the world can rent computers in them using services like Alibaba Cloud, Amazon Web Services, Google Cloud Platform, and Microsoft Azure.[20] Moreover, Alibaba, Amazon, and Google offer access to their custom processors for DNNs through their cloud service. The popularity of the public cloud is indicated by its annual growth in business by up to 50% since 2010 [Sch21]. Many believe the cloud's efficiencies in cost and energy mean that it is the ultimate future of all datacenters [Arm10, Sch21].

The next topic reminds us that reducing cost and energy consumption remains important no matter how green the cloud becomes.

### 4.7 Reducing the cost of training matters too

Though many have access to these relatively efficient compute resources and cloud companies may dramatically reduce their carbon footprint in the future, it's still important to reduce the economic *cost* of training. Saving money obviously matters to everyone, but expensive training of NLP models also makes this research style unattainable for many researchers[21,22]. This inequity of access to state-of-the-art models is another strong motivator, alongside environmental concerns, to incentivize the development of energy-efficient ML models that work as well as their computationally hungrier counterparts.

One issue that was difficult for us during our investigation was to put into perspective the 4 to 552 $tCO_2e$ from training of these NLP models, which the next subsection explores.

### 4.8 How does training a large NLP model compare to other activities?

Google Flights estimate for the emissions of a direct round trip of a whole passenger jet between San Francisco and New York is 180 $tCO_2e$ (see Table 2 and Appendix A). T5 training emissions are ~26%, Meena is 53%, Gshard-600B is ~2%, Switch Transformer is 32%, and GPT-3 is ~305% of such a round trip.

Another comparison point is to Bitcoin. Every purchase that transfers bitcoin currently costs ~700 KWh or ~0.3 $tCO_2e$, equivalent to the $CO_2e$ produced by ~750,000 credit card swipes. Bitcoin miners use custom chips that operate continuously 24/7 until they fail. Estimates of Bitcoin's impact for 2021 are ~78–121 TeraWatt-hours and ~37M–58M $tCO_2e$ [Cri21, Dig21]. Stated alternatively, ~70M people have Bitcoin wallets yet Google consumes 1/10th of Bitcoin's energy to provide services for billions of people, and all of Google's energy use is offset. If Bitcoin were a country, it would be in the top 30 in $CO_2e$; larger than Argentina, whose population is 45M. The estimated annual carbon footprint of Bitcoin mining this year is equivalent to roughly 200,000 to 300,000 whole passenger jet SF↔NY round trips.

In 2019 the world saw 39M flights and US airlines flew 925M passengers, which helps explain why air travel was responsible for 940 $MtCO_2$, or ~2.5% of the world's annual $CO_2$ in 2018 of 33B $tCO_2e$ [Rit20].

Finally, Google publishes its total energy consumption, and for 2019 it was 12.2 TeraWatt-hours [Goo20]. Row 18 of Table 4 shows the percentage that each NLP model training was of that total. Even if we assume all four of Google's large NLP models in Table 4 were trained in 2019, the total represents less than 0.005%. **The training of those four large NLP models is not a significant fraction of Google's energy consumption.**

---

[19] Improvement in energy intensity is expressed as energy use per compute instance. [Has20] goes on to say the cloud's increasing share of datacenters is causing a "notable improvement compared with recent annual efficiency gains in other major demand sectors (e.g., aviation and industry), which are an order of magnitude lower."
[20] There are not many cloud companies. With new technologies, initially only a few firms can practice the technology and they sell it to others, but these companies compete. There are many examples. Chemical technologies are in the hands of a relatively small number of companies; only six or seven institutions worldwide can refine crude oil; just a few firms can manufacture computer chips in the finest technology node (3–5 nm).
[21] To support the goal of making ML more inclusive, Google provides free access to a total of ~500 PetaFLOPS/second of TPU compute power to help ML researchers around the world participate in advancing the start of the art of ML.
[22] One possible unintended consequence of making training of a model less expensive is that more people will train the model and increase energy use, but that seems like a better risk than to continue using inefficient models.

Having spent 13 pages on the cost of large NLP models and neural architecture search, we conclude our discussion with three examples of the potential benefits of NLP models.

## 4.9 Are the benefits of NLP models worth the energy cost?

A recent example of a societal benefit of NLP is the COVID-19 Research Explorer, which helps scientists and researchers efficiently pore through articles for answers or evidence to COVID-19-related questions. It is powered by BERT, a Transformer-style model trained for the biomedical domain [Hal20].[23] Its training consumed ~2.8 MWh and produced 0.13 t$CO_2$e, about one-tenth of a SF-NY round trip by one passenger.[24]

A more widespread example is the use of BERT in search. English is the most popular language on the web. This use of BERT takes models that learn from improvements in English and applies them to other languages. In particular, BERT significantly improved featured snippets—short text summary at the top of Google research results—in languages like Hindi, Korean, and Portuguese.

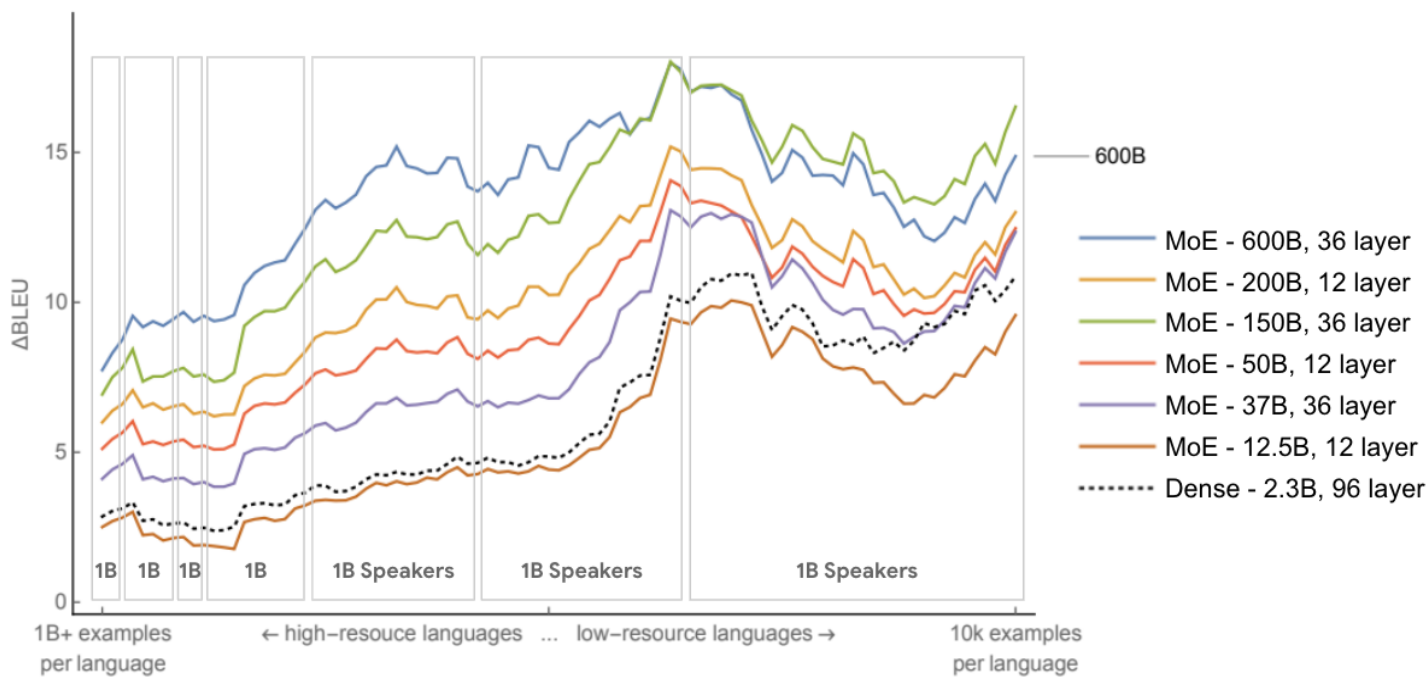

**Figure 7: Reproduction of Figure 6 from [Lep20] with annotations. Translation quality comparison of multilingual Mixture of Expert (MoE) Transformer models trained with GShard showing the increase in BLEU score versus a separate baseline Transformer model trained on each language pair for 100 languages to English. MoE models have large model capacity but are only partially activated for any given token. The source languages are grouped on the x-axis by the resources available for each language in billions of speakers, with languages like French and Spanish on the left (>1B examples) and languages like Sindhi and Yoruba on the right (<1M examples). The BLEU score improvements from larger models and multilingual training are high for all languages but are even higher for low-resource languages—the graph's right-hand side is higher than the left—so Yoruba translation quality benefits more than Spanish translation quality.**

A final example is the GShard multilingual translation model itself. Bender & Gebru *et al.* [Ben21] raise several legitimate issues in the development and use of large language models. Creating such models requires careful attention to issues of fairness and bias [Ben21, Gar19, Joh20, Kuc18, Mer19], but they also have the potential to benefit people everywhere. For example, our large scale translation models (M4) have already been used to translate billions of queries annually for each mid-to-low resource language[25] with 2B speakers globally for these languages. Figure 7, from the GShard paper [Lep20], shows substantial improvements for translation of 100 different languages to English. The blue line on the top in the left represents the 600B parameter multi-lingual translation MoE model of GShard. The dashed black line near the bottom is for a traditional dense DNN that is fully activated for every token. The dense DNN requires ~10X more computational resources to train than the 600B sparse MoE model, despite substantially lower translation quality. Figure 7 shows the larger MoE model, the larger the BLEU score gains were across all languages; the lines rarely cross. The 600B MoE model improves average quality +13.5 BLEU, 7.4 higher than the 2.3B dense model.

---

[23] Despite targeting a narrow audience of scientists, COVID explorer served 1000 queries per day at launch. It drew interest from Pfizer, Bristol Myers Squibb, AstraZeneca, Regeneron, British Medical Journal, European Food Safety Authority, and the National Institute of Health. Pfizer's Director of Global Medical Epidemiology used the tool daily; it led to Pfizer epidemiology research group to adapt the underlying ML models for systematic reviews and literature search.
[24] Training COVID Explorer took 6 days on 64 TPU v3s running in Oklahoma. It used ~2.8 MWh and 0.13 net t$CO_2$e.

GShard-600B's emissions (Table 4) are 4.3 t$CO_2$e —3.5 passenger SF-NY round trips—from consuming 24 MWh to train the model that could have 2B users; the amortized per-user $CO_2$e impact of model training would be less than the $CO_2$e impact of sending one text message[26].

## 5. Conclusion

Global climate change is a threat to economies, human health, and the environment, and the ML community needs to do its share to limit its carbon emissions.[27] We're thankful that papers like [Lac19, Str19, Sch20, Hen20] helped make the ML community aware of this important issue. Improving the energy efficiency of algorithms, datacenters, hardware, and software has long been a business priority for Google and other Cloud companies. For example, Gshard-600B operates much more efficiently than other large NLP models and ML accelerators are more efficient than off-the-shelf hardware.

As mentioned in the introduction, we make three suggestions for publications on compute intensive models that could eventually help reduce their $CO_2$e footprint: report energy consumed and $CO_2$e explicitly, ML conferences should reward improvements in efficiency as well as traditional metrics, and include the time and number of processors for training to help everyone understand its cost. We believe power will be included in upcoming MLPerf benchmarks, which is an important step in the right direction.

If the ML community working on computationally intensive models starts competing on training quality and carbon footprint rather than on accuracy alone, the most efficient datacenters and hardware might see the highest ML demand. If paired with publication incentives to improve emission metrics in addition to accuracy, we can imagine a virtuous cycle that slows the growth of the carbon footprint of ML by accelerating innovations in the efficiency and cost of algorithms, systems, hardware, datacenters, and carbon free energy.

## Acknowledgements

We wish to express our thanks to colleagues at Google and elsewhere who helped shape and improve this paper. Emma Strubell made several suggestions of ideas and organization of the paper, including suggesting adding data about the five large models. We thank Christopher Berner, Ilya Sutskever, OpenAI, and Microsoft for sharing information about GPT-3. Dmitry Lepikhin and Zongwei Zhou did a great deal of work to measure the performance and power of GPUs and TPUs in Google datacenters. Hallie Cramer, Anna Escuer, Elke Michlmayr, Kelli Wright, and Nick Zakrasek helped with the sections on energy and $CO_2$e emissions at Google. Tim Kraska suggested a revised organization of this paper. We thank Daniel Adiwardana, Gabriel Bender, Andrei Broder, Charina Chou, Jesse Dodge, Oren Etzioni, Orhan Firat, Ananya Ganesh, Robbie Gonzalez, David Grangier, Marsden Hanna, Urs Hölzle, Sheng Li, Sasha Luccioni, Preston McAfee, Andrew McCallum, Esteban Real, Stven Ross, Brennan Saeta, Roy Schwartz, Victor Schmidt, Ian Schneider, Aarush Selvan, Noah A. Smith, Zak Stone, Kate Weber, and Cliff Young for their help and feedback on the manuscript.

---

[25] In our setup for Figure 7, low resource languages have less than 1M training examples, mid resource languages have less than 10M training examples, and high resource languages have more than 1B training examples.
[26] An SMS message is 0.014 g of $CO_2$. That is larger than 24 MWh / 2B, which yields about 0.005 g of $CO_2$.
[27] We did not address the carbon footprint of ML in phones and other edge devices. It would be an excellent topic for another paper.

# Appendix A. Details of $CO_2$ Estimates for Four Large NLP Models in Tables 1 and 4

We describe below how we derived the values in Tables 1 and 4.

- *Datacenter Gross $CO_2e$/KWh (Table 1, row 4; Table 4, row 7):* The US Average is from [USE21]. For Google, we used the $CO_2e$ per KWh in the datacenter based at the time that the DNNs ran. (Here is a link for annual CFE% for Google Cloud.) For Microsoft, we use the 2020 US national average.
- *Datacenter Net $CO_2e$/KWh (Table 1, row 5; Table 4, row 8):* No change from above except for Google, where we used the net $CO_2e$ per KWh in the datacenter based on the 24/7 carbon-free energy methodology to estimate net carbon emissions at the time[28] that the DNNs ran (see Section 2.4 and Appendix B).
- *PUE (Table 1, row 6; Table 4, row 9)*: We use the Google datacenter PUE where the DNNs ran (published at https://www.google.com/about/datacenters/efficiency/). OpenAI told us that the PUE for the datacenter where GPT-3 ran was 1.10 [Sut21].
- *Measured Average Power (Table 1, row 9; Table 4, row 12)*: At Google we measured actual power usage rather than use Thermal Design Power (TDP), as TDP is a worst case for the chip. System power measurement includes the memory, fans, CPU host, network interface and so on, similar to the methodology of [Str19]. OpenAI measured V100s as running GPT-3 at 330W. GPUs can run on average closer to its TDP due to GPU's having Turbo Mode and Dynamic Voltage Frequency Scaling, not found in TPU v2/v3.
- *Measured Performance (Table 1, row 10; Table 4, row 13):* Profiling data was obtained via Google's internal performance analysis tool, Xprof. Measured FLOPs/s are calculated as the number of computed operations divided by execution time.
- *Number of Chips (Table 1, row 11; Table 4, row 14)*: We know the number of processors for the Google models. NVIDIA's press release about GPT-3 suggests OpenAI used 10,000 V100 GPUs for GPT-3.
- *Training time (Table 1, row 12; Table 4, row 15)*: We have the exact training time for Google DNNs. OpenAI published the total number of floating point operations to train their model: 3.14E+23 [Bro20]. OpenAI told us the V100 runs GPT-3 at 24.6 TeraFLOPS/sec [Sut21]. It takes ~14.8 days for 10,000 GPUs at 24.6 TeraFLOPS/sec to compute 3.14E+23 FLOPS. For the $CO_2e$ calculation, it doesn't actually matter whether it takes 2 weeks on 10,000 GPUs or 20 weeks on 1,000 GPUs, but we need one number for Table 4, so we used NVIDIA's suggestion of 10,000 GPUs.
- *Total Computation (Table 1, row 13; Table 4, row 16):* We calculate from measured performance, number of chips, and days to train (except for GPT-3, as OpenAI published the total FLOPS).
- *% of Google 2019 Energy Consumption. (Table 4, row 17):* For all models (even those not actually run in Google datacenters or not run in 2019), we calculate the percentage of Google's total energy consumption of 12.2 Terawatt-hours in 2019 [Goo20].
- *Ratio of round trips (Table 4, row 22).* To give perspective on the $CO_2e$ cost of training a model is compared to other activities, we show the $CO_2e$ of passenger jets. Google Flights calculated the average $CO_2$ emission for all the direct flights between San Francisco (SFO) and New York (JFK) in its database as 90.2t, so the average round trip is 180.4t. (This is for the whole plane, not just for one passenger.) Google Flights relies on this European Environmental Agency guidebook for these calculations and includes the minimum bounds for RF and NOx factor from Figure 6b in [Kär18].
- *% Carbon Free Energy (Table 1, row 17; Table 4, row 24).* Collected for when the models were run.

---

[28] All the 2020 datacenter measurements are provisional, awaiting final validation in May 2021

# Appendix B. Carbon Offset and 24/7 Carbon Free Energy

While energy consumption is relatively straightforward, policies to reduce carbon footprint are not. One reason is that they have as much to do about economics and accounting as they do about physics. This short appendix tries to clarify the distinction between conventional carbon offsets, Google's goal for 2030 of 24/7 Carbon Free Energy (CFE) for its global datacenters and campuses, and what it is doing in 2021 to set the groundwork for 2030. Readers interested in greater depth should take a look at [Ryo14, Goog16, Goo21].

Conventional carbon offsets try to create economic incentives to create projects that avoid or remove $CO_2e$. When pursuing the mitigation of carbon emissions from electricity production and consumption, a company can match their MWh of consumption with MWh of clean energy through certificates called *REC*s (*Renewable Energy Certificates*). The rules for accounting and compensation, are defined as part of the GHG Protocol, under Scope 2 for electricity. Under the current Scope 2 Guidance, 1MWh of energy used in July in, say, Georgia that produces carbon dioxide can be compensated by purchasing 1MWh of CFE in Montana in November. Typically, the period of accounting is a calendar year. Google achieved carbon neutrality using conventional carbon offsets starting in 2007.[29]

As part of the GHG Protocol, the World Resource Institute defines terms and economic mechanisms to ensure consistency of claims about carbon. They defined the following [Car21, Ryo14] (also see Figure 8):

- *Additionality*: $CO_2e$ reductions are *additional* if they would not have occurred in the absence of a market for offset credits. Additionality is essential for the quality of carbon offset credits—if their associated $CO_2e$ reductions are not additional, then purchasing offset credits in lieu of reducing your own emissions will make climate change worse.
- *The Grid*: The transmission and distribution system that connects generators and end-users.
- *Levelized Cost Of Energy (LCOE)*: The projected total system and operating costs divided by total KWh produced over the lifetime of the project or contract.
- *Power Purchase Agreement (PPA)*: A fixed-price contractual agreement to purchase a power plant's energy, typically calculated using LCOE.
- *Renewable Energy Certificate (REC)*[30]: A market-based instrument that represents the property rights to the environmental, social, and other non-power attributes of renewable electricity generation. The goal is a certificate that ensures the energy purchased is genuinely renewable and not double counted.

Google's target for 2030 is to go beyond the traditional Scope 2 rules to restrict both the location and the accounting period.

- Instead of anywhere in a continent, the CFE purchase should be on the same geographically local grid.
- Instead of the accounting period being one year, the accounting should be within the hour.

To achieve 100% 24/7 local CFE, grids would need to offer both real time accounting of the CFE fraction of the standard grid and the generating companies must offer more flexible options to allow consumers to pick CFE any time of the day, not just when the wind blows or when the sun shines. Ideally, grid operators and generating companies will deliver on that vision, and the standards will evolve to certify and quantify the 24/7 CFE approach. But we are not there yet.

Figure 8 helps explain what Google is doing today. Google signs long-term contracts as PPAs with renewable energy generating companies to try to cover Google's electricity consumption.[31] One benefit of long-term contracts is that they guarantee a reliable income stream for many years and therefore make such projects more easily financeable. To hit its 24/7 target, Google will continue to purchase clean energy from various sources such as energy storage and energy generation to ensure it has a clean energy supply at all 24 hours of the day, 7 days a week.

---

[29] In 2017, Google became the first major company to match 100% of its annual electricity use with renewable energy—purchasing as much clean energy as it consumed —which it has done for three consecutive years.
[30] RECs are more properly called *Energy Attribute Certificates*. Europe calls them *Guarantees of Origin* (*GOs*), not RECs.
[31] Google's more than 50 long-term contracts to purchase renewable energy resulted in more than $7 billion in new capital investment in renewable energy projects worldwide as of September 2019 [Goo20].

The percentage of CFE for a datacenter is reported ex-post, after load, production, and grid mix data are settled and made available to Google. With the current 24/7 CFE framework, when Google cannot get 100% CFE from the grid plus its clean energy contracts in a given hour, the shortfall counts against the goal. When the grid and renewable energy contracts overshoot in a given hour, Google doesn't get any extra credit for it, as the accounting period is reset every hour.[32] Since Google can estimate how much CFE is expected in a specific region based on the grid and its multi-year clean energy contract, it incentivizes programs to run in this region.[33]

Tables 1 and 4 show this distinction as *gross $CO_2e$* (energy from the grid) and the *net $CO_2e$* (after applying the 24/7 local renewable energy purchase from the long-term contracts). Since you can't label electrons, there is no guarantee that Google is using exactly the same clean energy that it paid for, but in our view the overall effect is the same.

Alas, Google's large models in Table 4 were run in the Georgia datacenter, where in the past there was no or little difference between gross and net $CO_2e$. Regions that have generator companies that can supply clean energy 24/7 and offer marketplaces that allow companies to acquire clean energy at any time of day will be more compelling to expand future growth of compute from a carbon impact perspective. A great example is Oklahoma, which allowed Google to average 95.6% net CFE for 2020. This is a case of where the grass actually is greener in Oklahoma than in Georgia. As mentioned above, in 2021 many new TPU v4 accelerators will be deployed in windy Oklahoma.

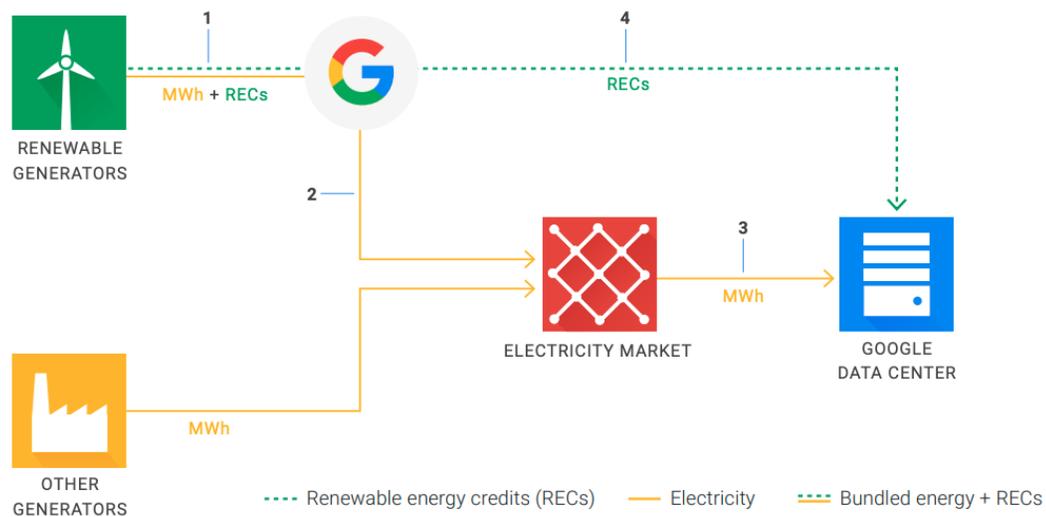

**Figure 8. This figure explains how fixed-floating swaps work for Renewable Energy Certificates (RECs). (Reproduced from [Goo16].) Instead of accounting over a full year at a mix of locations as in step 4, 24/7 CFE does the accounting separately for every hour in the year in the same single location.**

---

[32] Excess CFE from Google projects is used to support other grid load as well as incentivizing additional renewable development by demonstrating demand and driving down prices.
[33] Google even deployed a system in 2020 that shifts the timing of non-urgent compute tasks (like ML training) to when carbon-free power sources are most plentiful [Rad20]. Its next iteration will even move a task to a new datacenter.

## Appendix C. Details of a $CO_2e$ Estimate for NAS in an Average Datacenter

[Str19] estimates the $CO_2e$ for the neural architecture search (NAS) to find the more-efficient Evolved Transformer architecture done by [So19] at Google as 626,155 pounds (284 $tCO_2e$). The estimate in [Str19] was done for the hypothetical scenario of running the computation on P100 GPUs in the average U.S. datacenter with the average U.S. grid energy mix. The authors of this note represent a superset of the authors of [So19], and we agree that the information needed for an accurate estimate was scattered in several subsections in the So *et al*. paper, which makes it difficult to determine the actual $CO_2e$. This experience is one reason we suggest that ML conferences encourage future NLP papers that are computationally expensive to include a calculation of energy consumed and $CO_2e$ to make sure all the details are included, as it's difficult to determine them retrospectively, as we shall see.

NAS costs in [Str19] are derived from the NAS process described in section 5.2 of [So19]:

> "The search ran for 15K child models, requiring a total of 979M train steps. Over 13K models did not make it past the first hurdle, drastically reducing the resources required to view the 240 thousandth train step for top models, which would have cost 3.6B training steps for the same number of models without hurdles. After the search concluded, we then selected the top 20 models and trained them for the full 300K steps, each on a single TPU V.2 chip."

The projection of the So *et al*. NAS cost by Strubell *et al*. overestimates the actual Evolved Transformer search cost. Strubell *et al*. assumed each evaluation in the search is conducted using a large configuration: Transformer (Big) with batch size 32,768. However, So *et al*. actually used a small proxy configuration (Section 3.3 of [So19]) to reduce compute cost (and $CO_2e$). This proxy version used Transformer (Base) rather than Transformer (Big), reducing the cost/step by 2.3x. It also reduced the training batch size from 32,768 to 4,096 while keeping the number of training steps unchanged, reducing the cost/step by a further 8x.

As a result, the calculations below suggest that $CO_2e$ from the misunderstanding about the use of the smaller proxy task were overestimated by a factor of ~18.7:

Assume the Carbon Emission Estimation Method in [Str19]:
  $CO_2e$ = num_chips x num_train_steps x hours/train_steps x emission/chip_per_hour
  num_train_steps = 979,000,000  # From [So19]
  emission_per_chip_per_hour ~= 0.2855296 pounds $CO_2e$ # From [Str19] Table 3[34].

Estimation of Compute Cost in [Str19]:
  8 P100s for batch size 32,768 (packed version) from [Vas17] (4096 per GPU):
  num_chips = 8
  The Training speed of Transformer Big on P100 from [Vas17]:
  hours_per_train_steps = 84 hours / 300,000 = 0.00028 (Section 5.2 in [Vas17])
  $CO_2e$ = 8 * 979,000,000 * 0.00028 * 0.2855296 = **626,155 lbs (284 t)**

Estimation of Compute Cost if using GPUs of the Actual Setting Adopted in [So19]:
  1 P100 for batch size 32,768 / 8=4096 (Section 4.1 second paragraph in [So19]).
  num_chips = 1 (Section 4.3 in [So19], note that the actual search used one TPU v2 chip to fit the same batch size as one P100)
  Training speed of Transformer *Base* on P100 from [Vas17]:
  hours_per_train_steps = 12 hours / 100,000 = 0.00012 (Section 5.2 in [Vas17])
  $CO_2e$ = 1 * 979,000,000 * 0.00012 * 0.2855296 = **33,544 lbs (15.2 t)**

Appendix D shows a ~5X further reduction in $CO_2e$ by adjusting for the hardware and datacenter where the NAS occurred rather than for P100s in a hypothetical US average datacenter.

---

[34] In this calculation, emission_per_chip_per_hour = average power per chip (in Watts) * PUE * lbs $CO_2e$ per Watt.

## Appendix D. Details of a $CO_2e$ Estimate for Google's Actual NAS

To calculate the emissions of the actual NAS in [So19] at Google, where the search was actually performed, we must adjust by three more factors beyond the assumptions in Appendix C:

1. We use Google Georgia datacenter's PUE from the period in which the search computation was run (1.10 in Table 4) instead of the US average in 2018 (1.58).
2. Strubell *et al.* used the US average $CO_2$ per kilowatt hour (KWh) as calculated by the U.S. Environmental Protection Agency (EPA) of 0.423 kg per KWh in 2018. For Google, we use the Georgia datacenter's average $CO_2e$/KWh for the month when NAS was performed (0.431 $CO_2e$/KWh in Table 4).
3. So *et al.* used Google TPU v2 accelerators, not NVIDIA P100 GPUs as modeled in [Str19]. TPU v2s are much faster, so the search process takes 32,633 TPU v2 hours instead of 117,780 P100 hours. We measured the power when running the [So19] NAS computation on TPU v2, including the memory, fans, network interfaces, and the CPU host. The average power was 208 Watts. [Str19] estimated the power per P100 as 189 Watts[35]. The performance/Watt for NAS of TPU v2 improved ( 117,780 / 32,633 ) * ( 189 / 208 ) or 3.3X.

Our estimate of the actual NAS search that So *et al.* ran at Google after adjusting for the correct datacenter PUE, $CO_2e$/KWh, and hardware is (6.8 * 24 * 200 * 208 * 1.10 / 1000) * 0.431 / 1000 = 3.2 $tCO_2e$ (7096 lbs).[36] **This actual emissions value is 88X smaller than the incorrect estimate of the carbon emissions of this search found in Strubell *et al.*** If we reran the NAS search today on TPU v2s in Google's Iowa datacenter with 24/7 local, real time net $CO_2e$ reduction instead of Google's Georgia datacenter, it would drop from 3.2 $tCO_2e$ to 0.6 $tCO_2e$ (476X smaller). If we reran using newer TPUs, $tCO_2e$ would shrink further.

When, where, how, and on which hardware training occurs matters in addition to what DNN is trained, which is why it's best to include energy consumed and $CO_2e$ in a publication rather than relying on others to estimate it correctly afterwards.

---

[35] Strubell *et al.* used a mix of tools to estimate power for GPU, host CPU, and host memory at 189 Watts, which they used to estimate NAS. Our measurements for P100 are much higher in Table 4 for Transformer (Big) 296 Watts. We included everything in the rack like we do for TPUs, including TPU memory, top of rack switch, fans, power supplies, and so on. The two systems are running different implementations of the same problem and the CPU hosts are different. One issue might be that NVIDIA's power measurement tool used in [Str18] samples power once a minute, so there may be sampling issues.

[36] To put 3.2 net $tCO_2e$ into perspective, Table 1 and Appendix A use Google Flights to calculate the $CO_2e$ for the average direct round trip flights between SFO and JFK as 180.4t. The Boeing 767 that United Airlines flies on that route has 175 seats. Google Flights uses the historical average of 84.5% seat occupancy, yielding 1.2t of $CO_2e$ per passenger round trip. Thus, the $CO_2e$ equivalent of NAS is ~3 passengers taking a round trip between San Francisco and New York.